%% file: main.tex
\documentclass[11pt,letterpaper]{article}
\usepackage[letterpaper]{geometry}
\usepackage{emnlp2016}
\usepackage{times}
\usepackage{latexsym}
\usepackage{amsmath}
\usepackage{amssymb}
\usepackage{amsfonts}
\usepackage{multirow}
\usepackage{url}
\usepackage{color}
\usepackage{graphicx}
\usepackage{lipsum}

\emnlpfinalcopy
\def\emnlppaperid{860}

\newcommand{\depinv}{$^{-1}$}

\newcommand{\pair}{\langle H,~w\rangle}

\makeatother

\title{Relations such as Hypernymy: Identifying and Exploiting Hearst Patterns in Distributional Vectors for Lexical Entailment}

\author{Stephen Roller \\
  Department of Computer Science \\
  The University of Texas at Austin \\
  {\tt roller@cs.utexas.edu} \\\And
  Katrin Erk \\
  Department of Linguistics \\
  The University of Texas at Austin \\
  {\tt katrin.erk@mail.utexas.edu} \\}

\date{}

\begin{document}
\maketitle
\begin{abstract}
  We consider the task of predicting {\em lexical entailment} using
  distributional vectors. We perform a novel qualitative analysis of one
  existing model which was previously shown to only measure the prototypicality
  of word pairs. We find that the model strongly learns to identify hypernyms using
  {\em Hearst patterns}, which are well known to be predictive of lexical
  relations. We present a novel model which exploits this behavior as a method
  of feature extraction in an iterative procedure similar to Principal
  Component Analysis. Our model combines the extracted features with the
  strengths of other proposed models in the literature, and matches or
  outperforms prior work on multiple data sets.

\end{abstract}

\section{Introduction}
\input{intro}

\section{Background}
\label{sec:background}
\input{background}

\section{Data and Resources}
\label{sec:data}
\input{data}

\section{Motivating Analysis}
\label{sec:motivation}
\input{motivation}

\section{Proposed Model}
\label{sec:proposed}
\input{proposed}

\section{Experimental Setup and Evaluation}
\label{sec:exp}
\input{exp}

\section{Results}
\label{sec:results}
\input{results}

\input{analysis}

\section{Conclusion}
\input{conclusion}

\section*{Acknowledgments}
The authors would like to thank I. Beltagy, Vered Shwartz, Subhashini Venugopalan, and the
reviewers for their helpful comments and suggestions.  This research was
supported by the NSF grant IIS 1523637. We acknowledge the Texas Advanced
Computing Center for providing grid resources that contributed to these
results.

\clearpage
\bibliography{bib}
\bibliographystyle{emnlp2016}

\end{document}

%% file: intro.tex
As the field of Natural Language Processing has developed, more ambitious
semantic tasks are starting to be addressed, such as Question Answering (QA) and
Recognizing Textual Entailment (RTE). These systems often depend on the use of
lexical resources like WordNet in order to infer entailments
for individual words, but these resources are expensive to develop,
and always have limited coverage.

To address these issues, many works have considered on how lexical entailments can
be derived automatically using distributional semantics. Some focus
mostly on the use of unsupervised techniques, and study measures which emphasize
particular word relations \cite{baroni:2011:gems}. Many are based
on the Distributional Inclusion Hypothesis, which states that the contexts in
which a hypernym appears are a superset of its hyponyms' contexts
\cite{zhitomirsky-geffet:2005:acl,kotlerman:2010:nle}.
More recently, a great deal of work has pushed toward using supervised methods
\cite{baroni:2012:eacl,roller:2014:coling,weeds:2014:coling,levy:2015:naacl,kruszewski:2015:tacl},
varying by their experimental setup or proposed model.

Yet the literature disagrees about which models are strongest \cite{weeds:2014:coling,roller:2014:coling}, or even if they
work at all \cite{levy:2015:naacl}.
Indeed, \newcite{levy:2015:naacl} showed that two existing lexical entailment
models fail to account for similarity between the antecedent and
consequent, and conclude that such models are only learning to predict
{\em prototypicality}: that is, they predict that {\em cat} entails {\em animal}
because {\em animal} is usually entailed, and therefore will also predict
that {\em sofa} entails {\em animal}. Yet it remains unclear why such models
make for such strong baselines
\cite{weeds:2014:coling,kruszewski:2015:tacl,levy:2015:naacl}.

We present a novel qualitative analysis of one prototypicality
classifier, giving new insight into why prototypicality classifiers perform strongly in
the literature. We find the model overwhelmingly learns to identify hypernyms
using Hearst patterns available in the distributional space, like ``animals such as cats'' and ``animals
including cats.''  These patterns have long been used to identify lexical
relations \cite{hearst:1992:coling,snow:2004:nips}.

We propose a novel model which exploits this behavior as a method of feature
extraction, which we call {\em H-feature detectors}. Using an iterative procedure
similar to Principal Component Analysis, our model is able to extract and learn
using multiple H-feature detectors. Our model also integrates overall word
similarity and Distributional Inclusion, bringing together strengths of several
models in the literature. Our model matches or outperforms prior work on
multiple data sets. The code, data sets, and model predictions are made available
for future research.\footnote{\url{http://github.com/stephenroller/emnlp2016}}

%% file: background.tex
Research on lexical entailment using distributional semantics has now spanned
more than a decade, and has been approached using both unsupervised
\cite{weeds:2004:coling,kotlerman:2010:nle,lenci:2012:starsem,santus:2013:thesis}
and supervised techniques
\cite{baroni:2012:eacl,fu:2014:acl,roller:2014:coling,weeds:2014:coling,kruszewski:2015:tacl,levy:2015:naacl,turney:2015:nle,santus:2016:lrec}.
Most of the work in unsupervised methods is based on the Distributional
Inclusion Hypothesis \cite{weeds:2004:coling,zhitomirsky-geffet:2005:acl}, which states that the contexts
in which a hypernym appear should be a superset over its hyponyms' contexts.

This work focuses primarily on the supervised works in the literature.
Formally, we consider methods which treat lexical entailment as a supervised
classification problem, which take as input the distributional vectors for a
pair of words, $(H,~w)$, and predict on whether the antecedent $w$ entails the
consequent $H$.\footnote{We use the notation $w$ and $H$ for {\em word} and
{\em hypernym}. These variables refer to either the lexical items, or their
distributional vectors, depending on context.}

One of the earliest supervised approaches was Concat \cite{baroni:2012:eacl}.
In this work, the {\em concatenation} of the pair $\pair$ was used as input
to an off-the-shelf SVM classifier. At the time, it was very successful, but later
works noted that it had major problems with {\em lexical memorization}
\cite{roller:2014:coling,weeds:2014:coling,levy:2015:naacl}. That is, when
the training and test sets were carefully constructed to ensure they were
completely disjoint, it performed extremely poorly. Nonetheless, Concat is
continually used as a strong baseline in more recent work
\cite{kruszewski:2015:tacl}.

In response to these issues of lexical memorization, alternative models
were proposed. Of particular note are the {\em Diff}
\cite{fu:2014:acl,weeds:2014:coling} and {\em Asym} classifiers
\cite{roller:2014:coling}. The Diff model takes the vector difference
$H-w$ as input, while the Asym model uses both the vector difference and
the squared vector difference as input.
\newcite{weeds:2014:coling} found that Concat moderately outperformed Diff,
while \newcite{roller:2014:coling} found that Asym outperformed
Concat. Both Diff and Asym can also be seen as a form of supervised
Distributional Inclusion Hypothesis, with the vector difference being analogous
to the set-inclusion measures of some unsupervised techniques \cite{roller:2014:coling}.
All of these works focused exclusively on {\em hypernymy detection},
rather than the more general task of lexical entailment.

Recently, other works have begun to analyze Concat and Diff for their ability
to go beyond just hypernymy detection. \newcite{vylomova:2016:acl} take an
extensive look at Diff's ability to model a wide variety of lexical relations
and conclude it is generally robust, and \newcite{kruszewski:2015:tacl} have
success with a neural network model based on the Distributional Inclusion
Hypothesis.

On the other hand,
\newcite{levy:2015:naacl} analyze both Concat and Diff in their ability to
detect general lexical entailment on five data sets: two consisting of
only hypernymy, and three covering a wide variety of other entailing
word relations. They find that both Concat and
Diff fail, and analytically show that they are learning to predict the {\em
prototypicality} of the consequent $H$, rather than the relationship between the
antecedent and the consequent, and consider this a form of lexical memorization.
They propose a new model, Ksim, which addresses
their concerns, but lacks any notion of Distributional Inclusion.
In particular, they argue for directly including the
cosine similarity of $w$ and $H$ as a term in a custom SVM kernel, in
order to determine whether $w$ and $H$ are related all. Ultimately,
\newcite{levy:2015:naacl} conclude that distributional vectors may simply be
the wrong tool for the job.

%% file: data.tex
Prior work on lexical entailment relied on a variety of
data sets, each constructed in a different manner. We focus on
four different data sets, each of which has been used for evaluation in
prior work. Two data sets contain only hypernymy relations, and two consider
general lexical entailment.

Our first data set is {\bf LEDS}, the Lexical Entailment Data Set,
originally created by \newcite{baroni:2012:eacl}.
The data set contains 1385 hyponym-hypernym pairs extracted
directly from WordNet, forming a set of positive examples. Negative examples
were generated by randomly shuffling the original set of 1385 pairs. As such,
LEDS only contains examples of hypernymy and random relations.

Another major data set has been {\bf BLESS}, the \newcite{baroni:2011:gems}
Evaluation of Semantic Spaces. The data set contains annotations of word
relations for 200 unambiguous, concrete nouns from 17 broad
categories. Each noun is annotated with its co-hyponyms, meronyms, hypernym
and some random words. In this work, we treat hypernymy as positive,
and other relations as negative.

These two data sets form our hypernymy data sets, but we cannot overstate their
important differences: LEDS is balanced, while BLESS contains
mostly negative examples; negatives in BLESS include both random pairs {\em
and} pairs exhibiting other strong semantic relations, while LEDS only
contains random pairs. Furthermore, all of the negative examples in LEDS are
the same lexical items as the positive items, which has strong implications on
the prototypicality argument of \newcite{levy:2015:naacl}.

The next data set we consider is {\bf Medical} \cite{levy:2014:conll}. This
data set contains high quality annotations of subject-verb-object entailments
extracted from medical texts, and transformed into noun-noun entailments by
argument alignments. The data contains 12,600 annotations, but only 945
positive examples encompassing various relations like hypernymy, meronomy,
synonymy and contextonymy.\footnote{A term for
entailments that occur in some contexts, but do not cleanly fit in other
categories; e.g. {\em hospital} entails {\em doctor}.} This makes it
one of the most difficult data sets: it is both domain specific and highly
unbalanced.

The final data set we consider is {\bf TM14}, a variation on the SemEval 2012
Shared Task of identifying the degree to which word pairs exhibit various
relations. These relationships include a small amount of hypernymy, but also
many more uncommon relations (agent-object, cause-effect, time-activity, etc).
Relationships were binarized into (non-)entailing pairs by
\newcite{turney:2015:nle}.  The data set covers 2188 pairs, 1084 of which are
entailing.

These two entailment data sets also contain important differences, especially
in contrast to the hypernymy data sets. Neither contains any random negative
pairs, meaning general semantic similarity measures should be less useful; And
both exhibit a variety of non-hypernymy relations, which are less
strictly defined and more difficult to model.

\subsection{Distributional Vectors}

In all experiments, we use a standard, count-based, syntactic distributional
vector space.  We use a corpus composed of the concatenation of Gigaword,
Wikipedia, BNC and ukWaC. We preprocess the corpus using Stanford CoreNLP 3.5.2
\cite{chen:2014:emnlp} for tokenization, lemmatization, POS-tagging and
universal dependency parses. We compute a syntactic distributional space for
the 250k most frequent lemmas by counting their dependency neighbors across the
corpus. We use only the top 1M most frequent dependency attachments as
contexts.  We use CoreNLP's ``collapsed dependencies'', in which prepositional
dependencies are collapsed e.g.  ``go to the store'' emits the tuples
(go,~prep:to+store) and (store,~prep:to\depinv+go).  After collecting counts,
vectors are transformed using PPMI, SVD reduced to 300 dimensions, and
normalized to unit length. The use of collapsed dependencies is very important,
as we will see in Section~\ref{sec:motivation}, but other parameters are
reasonably robust.

%% file: motivation.tex
As discussed in Section~\ref{sec:background}, the Concat classifier is
a classifier trained on the concatenation of the word vectors, $\pair$.
As additional background, we first review the findings of
\newcite{levy:2015:naacl}, who showed that Concat trained using a linear
classifier is only able to capture
notions of {\em prototypicality}; that is, Concat guesses that ({\em animal}, {\em sofa})
is a positive example because {\em animal} looks like a hypernym.

Formally, a linear classifier like Logistic Regression or Linear SVM learns a
decision hyperplane represented by a vector $\hat p$.
Data points are compared to this plane with the
inner product: those above the plane (positive inner product) are classified as
entailing, and those below as non-entailing. Crucially, since the input
features are the concatenation of the pair vectors $\pair$, the hyperplane
$\hat p$ vector can be {\em decomposed} into separate $H$ and $w$ components.
Namely, if we rewrite the decision
plane $\hat p = \langle \hat H, \hat w\rangle$, we find that each pair
$\pair$ is classified using:
\begin{equation}\label{eq:proto}
\begin{aligned}
  & \hat p^{\top}\pair\\
  &= \langle \hat H, \hat w \rangle^{\top}\pair\\
  &= \hat H^{\top}H + \hat w^{\top}w.
\end{aligned}
\end{equation}
This analysis shows that, when the hyperplane $\hat p$ is evaluated on a novel
pair, it lacks any form of direct interaction between $H$ and $w$ like the
inner product $H^\top w$. Without any interaction terms, the Concat classifier
has no way of estimating the relationship {\em between} the two words, and
instead only makes predictions based on two independent terms, $\hat H$ and
$\hat w$, the {\em prototypicality vectors}.
Furthermore, the Diff
classifier can be analyzed in the same fashion and therefore has the same fatal
property.

We agree with this prototypicality interpretation, although we believe it is incomplete:
while it places a fundamental ceiling on the performance of these classifiers, it
does not explain {\em why} others have found them to persist as strong
baselines
\cite{weeds:2014:coling,roller:2014:coling,kruszewski:2015:tacl,vylomova:2016:acl}.
To approach this question, we consider a baseline Concat classifier trained
using a linear model. This classifier should most strongly exhibit the prototypicality
behavior according to Equation~\ref{eq:proto}, making it the best
choice for analysis. We first consider the most pessimistic hypothesis: is it
only learning to memorize which words are hypernyms at all?

We train the baseline Concat classifier using Logistic Regression on each of
the four data sets, and extract the vocabulary words which are most similar to
the $\hat H$ half of the learned hyperplane $\hat p$. If the classifier is only
learning to memorize the training data, we would expect items from the data to
dominate this list of closest vocabulary terms. Table~\ref{tab:wordsim}
gives the five most similar words to the learned hyperplane, with bold words
appearing directly in the data set.

\begin{table}[t]
\begin{center}
  \begin{footnotesize}
  \begin{tabular}{|llll|}
    \hline
    LEDS & BLESS & Medical & TM14\\
    \hline
     material       &      goods             &     item           &      sensitiveness          \\
     structure      &      lifeform          &     unlockable     &      tactility              \\
     object         & {\bf item}             &     succor         &      palate                 \\
     process        & {\bf equipment}        &     team-up        &      stiffness              \\
     activity       & {\bf herbivore}        &     non-essential  &      content                \\
    \hline
  \end{tabular}
  \end{footnotesize}
\end{center}
\caption{Most similar words to the prototype $\hat H$ learned by the Concat model. Bold items
appear in the data set.}
\label{tab:wordsim}
\end{table}

Interestingly, we notice there are very few bold words at all in the list.
In LEDS, we actually see some hypernyms of data set items that do not
even appear in the data set, and the Medical and TM14 words do not even appear
related to the content of the data sets. Similar results were also found for
Diff and Asym, and both when using Linear SVM and Logistic Regression. These
lists cannot explain the success of the prototypicality classifiers
in prior work. Instead, we propose an alternative interpretation of the
hyperplane: that of a feature detector for hypernyms, or an
{\em H-feature detector}.

\subsection{H-Feature Detectors}

Recall that distributional vectors are derived from a matrix $M$ containing
counts of how often words co-occur with the different syntactic contexts. This
co-occurrence matrix is factorized using Singular Value Decomposition,
producing both $W$, the ubiquitous word-embedding matrix, and $C$, the context-embedding
matrix \cite{levy:2014:nips}:
\begin{equation*}
  M \approx WC^{\top}
\end{equation*}
Since the word and context embeddings implicitly live in the same vector space
\cite{melamud:2015:vsm}, we can also compare Concat's hyperplane with the
context matrix $C$. Under this interpretation, the Concat model
does not learn what words are hypernyms, but rather what {\em contexts} or
{\em features} are indicative of hypernymy.
Table~\ref{tab:ctxsim} shows the syntactic contexts with the
highest cosine similarity to the $\hat H$ prototype for each of the different data sets.

\begin{table}[t]
\centering
\begin{small}
\begin{tabular}{|ll|}
  \hline
  LEDS & BLESS\\
  \hline
    nmod:such\_as+animal             &  nmod:such\_as+submarine           \\
    acl:relcl+identifiable           &  nmod:such\_as+ship                \\
    nmod:of\depinv+determine         &  nmod:such\_as+seal                \\
    nmod:of\depinv+categorisation    &  nmod:such\_as+plane               \\
    compound+many                    &  nmod:such\_as+rack                \\
    nmod:such\_as+pot                &  nmod:such\_as+rope                \\
  \hline
  Medical & TM14\\
  \hline
    nmod:such\_as+patch              &  amod+desire                       \\
    nmod:such\_as+skin               &  amod+heighten                     \\
    nmod:including+skin              &  nsubj\depinv+disparate            \\
    nmod:such\_as+tooth              &  nmod:such\_as+honey               \\
    nmod:such\_as+feather            &  nmod:with\depinv+body             \\
    nmod:including+finger            &  nsubj\depinv+unconstrained        \\
\hline
\end{tabular}
\end{small}
\caption{Most similar contexts to the prototype $\hat H$ learned by the Concat model.}
\label{tab:ctxsim}
\end{table}

This view of Concat as an H-feature detector produces a radically different perspective on
the classifier's hyperplane. Nearly all of the features learned take the form of
Hearst patterns \cite{hearst:1992:coling,snow:2004:nips}.
The most recognizable
and common pattern learned is the ``such as'' pattern, as in ``animals such as
cats''.  These patterns have been well known to be indicative of hypernymy
for over two decades. Other interesting patterns are the ``including'' pattern
(``animals including cats'') and ``many'' pattern (``many animals''). Although
we list only the six most similar context items for the data sets, we find
similar contexts continue to dominate the list for the next
30-50 items. Taken together, it is remarkable that the model identified these
patterns using only distributional vectors and only the positive/negative
example pairs. However, the reader should note these are not true Hearst
patterns: Hearst patterns explicitly relate a hypernym and hyponym using an
exact pattern match of a {\em single} co-occurrence. On the other hand, these
{\em H-features} are {\em aggregate indicators} of hypernymy across a large
corpus.

These learned features are much more interpretable than those found in the
analysis of prior work like \newcite{roller:2014:coling} and
\newcite{levy:2015:naacl}. \newcite{roller:2014:coling} found no signals of
H-features in their analysis of one classifier, but their model was focused
on {\em bag-of-words} distributional vectors, which perform significantly worse
on the task.
\newcite{levy:2015:naacl} also performed an analysis of lexical entailment
classifiers, and found weak signals like ``such'' and ``of'' appearing as
prominent contexts in their classifier, giving an early hint of H-feature
detectors, but not to such an overwhelming degree as we see in this work.
Critically, their analysis focused on a classifier trained on high-dimensional,
{\em sparse} vectors, rather than focusing on {\em context embeddings} as we
do.  By using these sparse vectors, their model was unable to generalize across
similar contexts. Additionally, their model did not make use of
{\em collapsed dependencies}, making features like ``such'' much weaker signals
of entailment and therefore less dominant during analysis.

Among these remarkable lists, the LEDS and TM14 data sets stand out for having
much fewer ``such as'' patterns compared to BLESS and Medical. The reason
for this is explained by the construction of the data sets: since LEDS
contains the same words used as both positive and negative examples, the
classifier has a hard time picking out clear signal. The TM14 data set, however,
does not contain any such negative examples.

We hypothesize the TM14 data set contains too many diverse
and mutually exclusive forms of lexical entailment, like
instrument-goal (e.g. ``honey'' $\rightarrow$ ``sweetness'').
To test this, we retrained the model with only hypernymy as positive examples, and
all other relations as negative.  We find that ``such as'' type patterns become top features,
but also some interesting data specific features, like ``retailer of [clothes]''.
Examining the data shows it contains many consumer goods, like
``beverage'' or ``clothes'', which explains these features.

%% file: proposed.tex
As we saw in the previous section, Concat only acts as a sort of H-feature
detector for whether $H$ is a prototypical hypernym, but does not actually
infer the relationship between $H$ and $w$. Nonetheless, this is powerful
behavior which should still be used in combination with the insights of other
models like Ksim and Asym. To this end, we propose a novel model which exploits
Concat's H-feature detector behavior, extends its modeling power, and adds two
other types of evidence proposed in the literature: overall similarity, and
distributional inclusion.

Our model works through an iterative procedure similar to Principal Component
Analysis (PCA). Each iteration repeatedly trains a Concat classifier under the
assumption that it acts as an H-feature detector, and then explicitly discards
this information from the distributional vectors. By training a new
H-feature detector on these modified distributional vectors, we can find {\em additional}
features indicative of entailment which were missed by the first classifier.
The entire procedure is iteratively repeated similar to how in Principal
Component Analysis, the second principal component is computed after the first
principal component has been removed from the data.

\begin{figure}
  \begin{center}
  \includegraphics[width=0.30\textwidth]{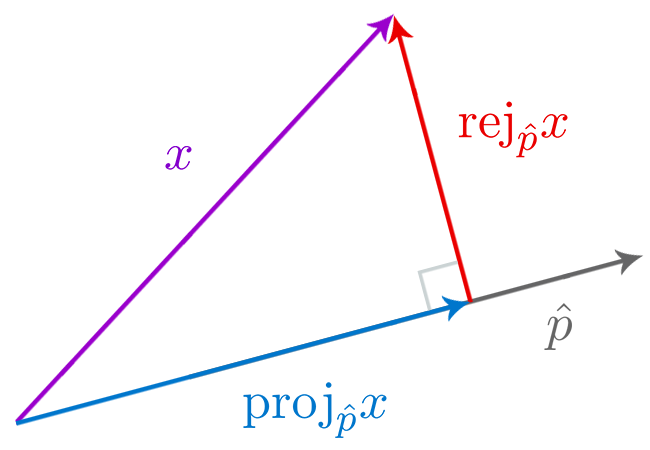}
\end{center}
\caption{A vector $\hat p$ is used to break $x$ into two orthogonal components,
its projection and the rejection over $\hat p$.}
\label{fig:vecproj}
\end{figure}

The main insight is that after training some H-feature detector using Concat,
we can {\em remove} this prototype from the distributional vectors through
the use of {\em vector projection}.
Formally, the vector projection of $x$ onto
a vector $\hat p$, $\text{proj}_{\hat p}(x)$ finds the {\em component} of $x$
which is in the direction of $\hat p$,
\begin{equation*}
  \text{proj}_{\hat p}(x) = \left(\frac{x^{\top}\hat p}{\|\hat p\|}\right)\hat p.
\end{equation*}
Figure~\ref{fig:vecproj} gives a geometric illustration of the vector
projection. If $x$ forms the hypotenuse of a right
triangle, $\text{proj}_{\hat p}(x)$ forms a leg of the triangle. This also
gives rise to the {\em vector rejection}, which is the vector forming the third
leg of the triangle. The vector rejection is orthogonal to the projection, and
intuitively, is the original vector after the projection has been removed:
\begin{equation*}
  \text{rej}_{\hat p}(x) = x - \text{proj}_{\hat p}(x).
\end{equation*}

Using the vector rejection, we take a learned H-feature detector $\hat p$,
and discard these features from each of the word vectors. That is, for every data
point $\langle H, w\rangle$, we replace it by its vector rejection and rescale
it to unit magnitude:
\begin{align*}
  H_{i+1} & = \text{rej}_{\hat p}(H) / \|\text{rej}_{\hat p}(H)\|\\
  w_{i+1} & = \text{rej}_{\hat p}(w) / \|\text{rej}_{\hat p}(w)\|
\end{align*}
A new classifier trained on the $\langle H_{i+1}, w_{i+1}\rangle$ data {\em must} now learn
a different decision plane than $\hat p$, as $\hat p$ is no longer present
in any data points. This repetition of the procedure is roughly analogous to
learning the second principal component of the data; we wish to classify the
pairs without using any information learned from the previous iteration.

This second classifier {\em must} perform strictly worse than the original,
otherwise the first classifier would have learned this second hyperplane.
Nonetheless, it will be able to learn {\em new} H-feature detectors
which the original classifier was unable to capture. By repeating this process,
we can find several H-feature detectors, $\hat p_1, \ldots, \hat p_n$. Although
the first, $\hat p_1$ is the best possible {\em single} H-feature detector,
each additional H-feature detector increases the model's representational power
(albeit with diminishing returns).

This procedure alone does not address the main concern of \newcite{levy:2015:naacl}:
that these linear classifiers never actually model any connection between $H$ and $w$.
To address this, we explicitly {\em compare} $H$ and $w$
by extracting additional information about how $H$ and $w$ interact with respect
to each of the H-feature detectors. This additional information is then used to
train one final classifier which makes the final prediction.

Concretely, in each iteration $i$ of the procedure, we generate a four-valued
feature vector
$F_i$, based on the H-feature detector $\hat p_i$. Each
feature vector contains (1) the similarity of $H_i$ and $w_i$ (before projection);
(2) the feature
$\hat p_i$ applied to $H_i$; (3) the H-feature detector $\hat p_i$ applied to $w_i$; and
(4) the difference of 2 and 3.
\begin{align*}
  & F_i(\langle H_i, w_i\rangle, \hat p_i)\\
  & \qquad = \langle H_i^{\top}w_i, H_i^{\top}\hat p_i, w_i^{\top}\hat p_i, (H_i - w_i)^{\top}\hat p_i\rangle
\end{align*}
These four ``meta''-features capture all the benefits of the
H-feature detector (slots 2 and 3), while still addressing Concat's issues with
similarity arguments (slot 1) {\em and} distributional inclusion (slot 4).
The final feature's relation to the DIH comes from the observation of
\newcite{roller:2014:coling} that the vector difference intuitively captures
whether the hypernym {\em includes} the hyponym.

The union of all the feature vectors $F_1, \ldots, F_n$ from repeated iteration
form a $4n$-dimensional feature vector which we use as input to one final
classifier which makes the ultimate decision.  This classifier is trained on
the same training data as each of the individual H-feature detectors, so
our iterative procedure acts {\em only} as a method of feature
extraction.

For our final classifier, we use an SVM with an RBF-kernel, though decision
trees and other nonlinear classifiers also perform reasonably well. The
nonlinear final classifier can be understood as doing a form of logical
reasoning about the four slots: ``animal'' is a hypernym of ``cat'' because (1)
they are similar words where (2) animal looks like a hypernym, but (3) cat does
not, and (4) some ``animal'' contexts are not good ``cat'' contexts.

%% file: exp.tex
In our experiments, we use a variation of 20-fold cross validation which
accounts for lexical overlap. To simplify explanation, we first explain how we
generate splits for training/testing, and then afterwards introduce validation
methodology.

We first pool all the words from the antecedent (LHS)
side of the data into a set, and split these lexical items into 20 distinct
cross-validation folds. For each fold $F_i$, we then use all pairs $(w, H)$ where
$w\in F_i$ as the test set pairs. That is, if ``car'' is in the test set fold,
then ``car $\rightarrow$ vehicle'' and ``car $\nrightarrow$ truck" will appear
as test set pairs. The training set will then be {\em every pair} which does
not contain {\em any} overlap with the test set; e.g. the training
set will be all pairs which do not contain ``car'', ``truck'' or ``vehicle''
as either the antecedent or consequent. This ensures that both (1) there
is zero lexical overlap between training and testing and (2) every pair is
used as an item in a test fold exactly once. One quirk of this setup is
that all test sets are approximately the same size, but training sizes
vary dramatically.

This setup differs from those of previous works like
\newcite{kruszewski:2015:tacl} and \newcite{levy:2015:naacl}, who both use
single, fixed train/test/val sets without lexical overlap. We find our setup
has several advantages over fixed sets. First, we find there can be
considerable variance if the train/test set is regenerated with a different
random seed, indicating that multiple trials are necessary. Second, fixed
setups consistently discard roughly half the data as ineligible for either
training or test, as lexical items appear in many pairs. Our CV-like setup
allows us to evaluate performance over every item in the data set exactly once,
making a much more efficient and representative use of the original data set.

Our performance metric is F1 score. This is more representative than
accuracy, as most of the data sets are heavily unbalanced. We report the mean
F1 scores across all cross validation folds.

\subsection{Hyperparameter Optimization}

In order to handle hyperparameter selection, we actually generate the test set
using fold $i$, and use fold $i-1$ as a validation set (removing pairs which
would overlap with test), and the remaining 18 folds as training (removing
pairs which would overlap with test {\em or} validation). We select
hyperparameters using grid search. For all models, we optimize over the regularization parameter
$C \in \{10^{-4}, 10^{-3}, \ldots, 10^4\}$, and for our proposed model, the
number of iterations $n \in \{1, \ldots, 6\}$. All other hyperparameters
are left as defaults provided by Scikit-Learn \cite{pedregosa:2013:jmlr},
except for using balanced class weights. Without balanced class weights,
several of the baseline models learn degenerate functions (e.g. always guess non-entailing).

%% file: results.tex
\begin{table}
\centering
\begin{small}
\begin{tabular}{|l|rrrr|}
  \hline
  Model            &      LEDS   &      BLESS  &      Medical  &      TM14   \\
  \hline
  \hline
  \multicolumn{5}{|c|}{Linear Models}\\
  \hline
  Cosine           &      .787   &      .208   &      .168     &      .676   \\
  Concat           &      .794   &      .612   &      .218     &      .693   \\
  Diff             &      .805   &      .440   &      .195     &      .665   \\
  Asym             &      .865   &      .510   &      .210     &      .671   \\
  Concat+Diff      &      .801   &      .604   &      .224     &      .703   \\
  Concat+Asym      &      .843   &  {\bf.631}  &      .240     &      .701   \\
  \hline
  \multicolumn{5}{|c|}{Nonlinear Models}\\
  \hline
  RBF              &      .779   &      .574   &      .215     &      .705   \\
  Ksim             &      .893   &      .488   &      .224     &  {\bf.707}  \\
  Our model        &  {\bf.901}  &  {\bf.631}  &  {\bf.260}    &      .697   \\
  \hline
\end{tabular}
\end{small}
\caption{Mean F1 scores for each model and data set.}
\label{tab:results}
\end{table}

We compare our proposed model to several existing and alternative baselines
from the literature. Namely, we include a baseline Cosine
classifier, which only learns a threshold which maximizes F1 score on the
training set; three linear models of prior work, Concat, Diff and Asym; and the
RBF and Ksim models found to be successful in
\newcite{kruszewski:2015:tacl} and \newcite{levy:2015:naacl}. We also include two additional
novel baselines, Concat+Diff and Concat+Asym, which add a notion of
Distributional Inclusion into the Concat baseline, but are still linear models.
We cannot include baselines like Ksim+Asym, because Ksim is based on a custom
SVM kernel which is not amenable to combinations.

Table~\ref{tab:results} the results across all four data sets for all of the
listed models. Our proposed model improves significantly\footnote{Bootstrap test, $p<.01$.} over Concat in the LEDS,
BLESS and Medical data sets, indicating the benefits of combining these aspects of
similarity and distributional inclusion with the H-feature detectors of Concat.
The Concat+Asym classifier also improves over the Concat baseline, further
emphasizing these benefits. Our model performs approximately the same as Ksim
on the LEDS and TM14 data sets (no significant difference),
while significantly outperforming it on BLESS and Medical data sets.

\subsection{Ablation Experiments}
\begin{table}
\centering
\begin{small}
\begin{tabular}{|l|rrrr|}
  \hline
  Model         &      LEDS   &      BLESS  &      Medical  &      TM14   \\
  \hline
  No Similarity &      .099   &      .061   &      .034     &      .003   \\
  No Detectors  &     -.008   &      .136   &      .018     &      .028   \\
  No Inclusion  &      .010   &      .031   &      .014     &      .001   \\
  \hline
\end{tabular}
\end{small}
\caption{Absolute decrease in mean F1 on the development sets with the
different feature types ablated. Higher numbers indicate greater feature
importance.}

\label{tab:ablation}
\end{table}

In order to evaluate how important each of the various $F$ features are to the
model, we also performed an ablation experiment where the classifier is {\em
not} given the similarity (slot 1), prototype H-feature detectors (slots 2 and
3) or the inclusion features (slot 4). To evaluate the importance of these
features, we fix the regularization parameter at $C = 1$, and train all
ablated classifiers on each training fold with number of iterations
$n = {1, \ldots, 6}$. Table~\ref{tab:ablation} shows the decrease (absolute difference) in performance
between the full and ablated models on the development sets, so higher numbers indicate greater
feature importance.

We find the similarity feature is extremely important in the LEDS, BLESS and
Medical data sets, therefore reinforcing the findings of
\newcite{levy:2015:naacl}. The similarity feature is especially important in
the LEDS and BLESS data sets, where negative examples include many random
pairs. The detector features are moderately important for the Medical and TM14 data sets,
and critically important on BLESS, where we found the strongest evidence
of Hearst patterns in the H-feature detectors.
Surprisingly, the detector features are moderately {\em detrimental} on the
LEDS data set, though this can also be understood in the data set's construction:
since the negative examples are randomly shuffled positive examples, the
{\em same} detector signal will appear in both positive and negative examples.
Finally, we find the model performs somewhat robustly without the inclusion
feature, but still is moderately impactful on three of the four data sets,
lending further evidence to the Distributional Inclusion Hypothesis.
In general, we find all three components are valuable sources of information
for identifying hypernymy and lexical entailment.

%% file: analysis.tex
\subsection{Analysis by Number of Iterations}

\begin{figure*}
  \begin{center}
  \includegraphics[width=1.00\textwidth]{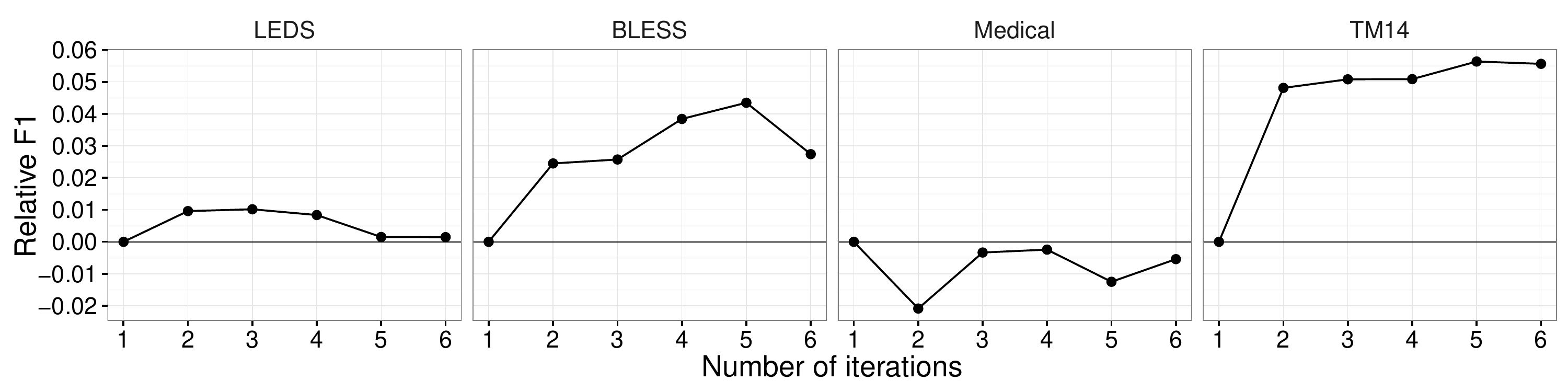}
\end{center}
\caption{Performance of model on development folds by number of iterations. Plots
show the improvement (absolute difference) in mean F1 over the model fixed at one
iteration.}
\label{fig:byiteration}
\end{figure*}

\begin{table*}
  \begin{center}
  \begin{small}
  \begin{tabular}{|llll|}
    \hline
    Iteration 1 & Iteration 2 & Iteration 3 & Iteration 4\\
    \hline
nmod:such\_as+submarine        &nmod:including+animal          &amod+free-swimming             &advcl+crown                         \\
nmod:such\_as+ship             &nmod:including+snail           &nmod:including\depinv+thing    &advcl+victorious                    \\
nmod:such\_as+seal             &nmod:including+insect          &nsubj\depinv+scarcer           &nsubj+eaters                        \\
nmod:such\_as+plane            &nmod:such\_as+crustacean       &nsubj\depinv+pupate            &nsubj+kaine                         \\
nmod:such\_as+rack             &nmod:such\_as+mollusc          &nmod:such\_as+mollusc          &nmod:at+finale                      \\
nmod:such\_as+rope             &nmod:such\_as+insect           &nmod:of\depinv+value           &nsubj+gowen                         \\
nmod:such\_as+box              &nmod:such\_as+animal           &nmod:as\depinv+exhibit         &nsubj+pillman                       \\
    \hline
  \end{tabular}
\end{small}
\end{center}
\caption{Most similar contexts to the H-feature detector for each iteration
of the PCA-like procedure. This model was trained on all data of BLESS. The first and second
iterations contain clear Hearst patterns, while the third and fourth contain
some data-specific and non-obvious signals.}
\label{tab:multiter}
\end{table*}

In order to evaluate how the iterative feature extraction affects model
performance, we fix the regularization parameter at $C = 1$, and train our
model fixing the number of iterations to $n = \{1, \ldots, 6\}$.
We then measure the mean F1 score across the development folds and compare to a
baseline which uses only one iteration. Figure~\ref{fig:byiteration} shows
these results across all four data sets, with the 0 line set at performance of
the $n = 1$ baseline. Models above 0 benefit from the additional
iterations, while models below do not.

In the figure, we see that the iterative procedure moderately improves
performance LEDS, while greatly improving the scores of BLESS and TM14, but
on the medical data set, additional iterations actually hurt performance.
The differing curves indicate that the optimal number of iterations is very
data set specific, and provides differing amounts of improvement, and therefore
should be tuned carefully. The LEDS and BLESS curves indicate a sort of
``sweet spot'' behavior, where further iterations degrade performance.

To gain some additional insight into what is captured by the various iterations
of the feature extraction procedure, we repeat the procedure from
Section~\ref{sec:motivation}: we train our model on the entire BLESS data set
using a fixed four iterations and regularization parameter. For each iteration,
we compare its learned H-feature detector to the context embeddings, and report
the most similar contexts for each iteration in Table~\ref{tab:multiter}.

The first iteration is
identical to the one in Table~\ref{tab:ctxsim}, as expected. The second
iteration includes many H-features not picked up by the first iteration,
mostly those of the form ``X including Y''. The third iteration picks up some
data set specific signal, like ``free-swimming [animal]'' and
``value of [computer]'', and so on. By the fourth iteration, the features
no longer exhibit any obvious Hearst patterns, perhaps exceeding the sweet
spot we observed in Figure~\ref{fig:byiteration}.
Nonetheless, we see how multiple iterations of the procedure allows our model
to capture many more useful features than a single Concat classifier on its
own.

%% file: conclusion.tex
We considered the task of detecting lexical entailment using distributional
vectors of word meaning.
Motivated by the fact that the Concat classifier acts as a
strong baseline in the literature, we proposed a novel interpretation of the
model's hyperplane. We found the Concat classifier overwhelmingly acted as a
feature detector which automatically identifies Hearst Patterns in the
distributional vectors.

We proposed a novel model that embraces these H-feature detectors fully, and extends
their modeling power through an iterative procedure similar to Principal
Component Analysis. In each iteration of the procedure, an H-feature detector
is learned, and then removed from the data, allowing us to identify several
different kinds of Hearst Patterns in the data. Our final model combines these
H-feature detectors with measurements of general similarity and Distributional
Inclusion, in order to integrate the strengths of different models in prior
work. Our model matches or exceeds the performance of prior work, both on
hypernymy detection and general lexical entailment.